\documentclass{article}
\usepackage{nips13submit_e,times}

\usepackage{natbib}
\usepackage{graphicx}
\usepackage{amsmath,amssymb,latexsym}
\usepackage{multirow}
\usepackage[applemac]{inputenc}

\input{def.set}

\usepackage{url}
\usepackage{amsmath}
\usepackage{amsthm}
\usepackage{amssymb}
\usepackage{graphicx}
\usepackage{xspace}
\usepackage{tabularx}
\usepackage{multicol}
\usepackage{multirow}
\usepackage{url}
\usepackage{wrapfig}
\usepackage[normalem]{ulem}
\usepackage{xcolor}

\newcommand{\RR}[0]{\mathbb{R}}

\newcommand{\NN}[0]{\ensuremath{\text{NN}}}
\newcommand{\TT}[0]{\ensuremath{\text{\btheta}}}

\newcommand{\UNKS}[1]{\ensuremath{\left<\text{UNK}\right>_{#1}}}
\newcommand{\NSym}[1]{\ensuremath{\left<\text{N}\right>_{#1}}}
\newcommand{\SSym}[1]{\ensuremath{\left<\text{S}\right>_{#1}}}
\newcommand{\CSym}[1]{\ensuremath{\left<\text{C}\right>_{#1}}}

\newcommand{\ola}{\overleftarrow}
\newcommand{\ora}{\overrightarrow}

\title{Context-Dependent Word Representation for Neural Machine Translation}

\author{Heeyoul Choi \\
Samsung Electronics\\
University of Montreal\\
\texttt{heeyoul@gmail.com}
 \And Kyunghyun Cho \\
New York University\\
\texttt{ kyunghyun.cho@nyu.edu}\\
\ \And Yoshua Bengio\\
University of Montreal\\
CIFAR Senior Fellow\\
\texttt{yoshua.bengio@umontreal.ca}}

\nipsfinalcopy % Uncomment for camera-ready version

\begin{document}
\maketitle

\begin{abstract}
    We first observe a potential weakness of continuous vector representations
    of symbols in neural machine translation. That is, the continuous vector
    representation, or a word embedding vector, of a symbol encodes multiple
    dimensions of similarity, equivalent to encoding more than one meaning of
    the word. This has the consequence that the encoder and decoder recurrent
    networks in neural machine translation need to spend substantial amount of
    their capacity in disambiguating source and target words based on the
    context which is defined by a source sentence. Based on this observation, in
    this paper we propose to contextualize the word embedding vectors using a
    nonlinear bag-of-words representation of the source sentence. Additionally,
    we propose to represent special tokens (such as numbers, proper nouns and
    acronyms) with typed symbols to facilitate translating those words that are
    not well-suited to be translated via continuous vectors. The experiments on
    En-Fr and En-De reveal that the proposed approaches of contextualization and
    symbolization improves the translation quality of neural machine translation
    systems significantly.
\end{abstract}

\section{Introduction}

Neural machine translation is a recently proposed paradigm in machine
translation, which is often entirely built as a single neural
network~\cite{kalchbrenner2013recurrent,Sutskever2014,Bahdanau2015}. The neural
machine translation system, which often consists of an encoder and decoder,
projects and manipulates a source sequence of discrete linguistic symbols
(source sentence) in a continuous vector space, and decodes a target sequence of
symbols (target sentence or translation.) This is contrary to the conventional
machine translation systems, such as phrase-based statistical machine
translation~\cite{koehn2003statistical}, which work directly at the discrete
symbol level. 

In more detail, the first step of any neural machine translation system is to
convert each atomic symbol into a corresponding continuous vector, which is
often called as a word embedding. This step is done for each source word
independently of the other words and results in a source sequence of word
embeddings. The encoder network, which is often implemented as a recurrent
neural network, encodes this source sentence either into a single context
vector~\cite{Sutskever2014,cho2014learning} or into a sequence of context 
vectors~\cite{kalchbrenner2013recurrent,Bahdanau2015}. 

The decoder network, again a recurrent neural network, generates a translation
word-by-word while being conditioned on the context representation of the source
sentence. At each step of generation in the decoder, the internal hidden state
of the decoder is updated first. The dot product between this hidden state and
the output word embedding vector of each word in the target vocabulary is
computed and normalized across all the target words, resulting in a probability
distribution over the target vocabulary. A target word is selected based on this
distribution, and the whole process is recursively repeated until the
end-of-sequence symbol is generated.

Among different variants of neural machine translation, the attention-based
approach \cite{Bahdanau2015} has recently become {\it de facto} standard. It has
been found to perform comparably to or better than the existing phrase-based
statistical systems in many language pairs including En-Fr~\cite{Jean2015},
En-De~\cite{Jean2015,jean2015montreal,luong2015effective},
En-Cs~\cite{jean2015montreal}, and En-Zh~\cite{Shen2015}. Much of these recent
improvements have been made by tackling, e.g., the attention mechanism (which is
central to the attention-based neural translation system) and the
computational issues arising from having a large target vocabulary. 

Unlike these recent works, we focus on source- and target-side word embedding
vectors in this paper. More specifically, we first notice that the
transformation from and to high-dimensional word embedding vectors is done for
each word largely independent of each other. We conjecture that only a few axes
in this high-dimensional space are relevant given a source sentence and that we
can remove much of the ambiguity in the choice of words by restricting, or
turning off, most of the irrelevant dimensions. We propose to achieve this
automated way to turn off some dimensions of word embeddings by {\it
contextualizing} a word embedding vector.

In addition to the proposed contextualization of both source and target word
embedding vectors, we propose to extend the unknown token replacement technique
proposed in \cite{Luong2015} to multiple token types. This extension, to which
we refer as {\it symbolization}, introduces multiple meta-tokens such as the
number token and the proper name token in addition to the unknown-word token. 
This symbolization effectively remaps rare tokens into more frequent
meta-tokens and thereby results in improved translation quality.

We extensively evaluate the proposed contextualization and symbolization on two
language pairs--En-Fr and En-De-- with the attention-based neural machine
translation model. The experiments reveal that the contextualization and
symbolization each improves the translation quality by 2 BLEU scores, and
together by 4 BLEU points on En-Fr. On En-De, they result in 1--2 BLEU score
improvements each, and together 2.5 BLEU score increase.

\section{Background: Neural Machine Translation}

In this section, we give a brief overview of neural machine translation.
More specifically, we describe the attention-based neural machine translation
\cite{Bahdanau2015}
which will be used in the experiments later. However, we note that the proposed
contextualization and symbolization techniques are generally applicable to any
other type of neural machine translation systems such as the
sequence-to-sequence model \cite{Sutskever2014}.

The attention-based neural machine translation system computes a conditional
distribution over translations given a source sentence
$X=(w^x_1, w^x_2, \ldots, w^x_T)$: 
\begin{align*}
    p(Y=(w^y_1, w^y_2, \ldots, w^y_{T'}) | X).
\end{align*}
This is done by a neural network that consists of an encoder, a decoder and the
attention mechanism. 

The encoder is often implemented as a bidirectional recurrent neural network
that reads the source sentence word-by-word. Before being read by the encoder,
each source word $w^x_t \in V$ is projected onto a continuous vector space:
\begin{align}
    \label{eq:word_emb_x}
    \bx_t = \bE^x \bold{1} (w^x_t),
\end{align}
where $\bold{1}(w^x_t)$ is a one-hot vector defined as
\begin{align}
    \label{eq:one_hot}
    \bold{1}(w^x_t)_j = \left\{ 
        \begin{array}{l l}
            1,&\text{ if }j=w^x_t \\
            0,&\text{ otherwise}
        \end{array}
        \right..
\end{align}
$\bE_x \in \RR^{E \times |V|} $ is a source word embedding matrix, where $E$ and
$|V|$ are the word embedding dimension and the vocabulary size, respectively. 

The resulting sequence of word embedding vectors is then read by the
bidirectional encoder recurrent network which consists of forward and reverse
recurrent networks. The forward recurrent network reads the sequence in the
left-to-right order:
\begin{align*}
    \ora{\bh}_t = \ora{\phi}(\ora{\bh}_{t-1}, \bx_t),
\end{align*}
while the reverse network reads it right-to-left:
\begin{align*}
    \ola{\bh}_t = \ola{\phi}(\ola{\bh}_{t+1}, \bx_t),
\end{align*}
where the initial hidden states $\ora{\bh}_0$ and $\ola{\bh}_{T+1}$ are
initialized as all-zero vectors.  The hidden states from the forward and reverse
recurrent networks are concatenated at each time step $t$ to form an annotation
vector $\bh$:
\mbox{$ \bh_t = \left[ \ora{\bh}_t; \ola{\bh}_t \right].  $}
This concatenation results in a context $C$ that is a tuple of annotation vectors:
\mbox{$ C = \left\{ \bh_1, \bh_2, \ldots, \bh_T \right\}.  $}
The recurrent activation functions $\ora{\phi}$ and $\ola{\phi}$ are in most
cases either long short-term memory units (LSTM, \cite{Hochreiter1997}) or gated
recurrent units (GRU, \cite{Cho2014}.) 

The decoder consists of two sub-components--a recurrent network and the attention
mechanism. The recurrent network in the decoder is a unidirectional language
model, which computes the conditional distribution over the next target word
given all the previous target words and the source sentence:
\[
    p(w^y_{t'}|w^y_{<t'}, X)
\]

The decoder recurrent network maintains an internal hidden state $\bz_{t'}$. At
each time step $t'$, it first uses the attention mechanism to select, or weight,
the annotation vectors in the context tuple $C$. The attention mechanism, which is
a feedforward neural network, takes as input both the previous decoder hidden
state, and one of the annotation vectors,
and returns a relevant score $e_{t' ,t}$:
\begin{align*}
    e_{t',t} = f_{\text{ATT}}(\bz_{t'-1}, \bh_t).
\end{align*}
These relevance scores are normalized to be positive and sum to 1:\footnote{
    For more variants of the attention mechanism, we refer the readers to
    \cite{luong2015effective}.
}
\begin{align}
    \label{eq:att_weight}
    \alpha_{t', t} = \frac{\exp(e_{t',t})}{\sum_{k=1}^T \exp(e_{t',k})}.
\end{align}
We use the normalized scores to compute the weighted sum of the annotation
vectors
\[
    \bc_{t'} = \sum_{t=1}^T \alpha_{t', t} \bh_t
\]
which will be used by the decoder recurrent network to update its own hidden
state by
\begin{align*}
    \bz_{t'} = \phi_z (\bz_{t'-1}, \by_{t'-1}, \bc_{t'}).
\end{align*}
Similarly to the encoder, $\phi_z$ is implemented as either an LSTM or GRU.
$\by_{t'-1}$ is a target-side word embedding vector computed by
\begin{align}
    \label{eq:word_emb_y}
    \by_{t'-1} = \bE^y \bold{1} (w^y_{t'-1}),
\end{align}
similarly to Eq.~\eqref{eq:word_emb_x}.

The probability of each word $i$ in the target vocabulary $V'$ is computed by
\begin{align*}
    p(w^y_{t'}=i | w^y_{<t'}, X) \propto \exp\left( 
        \bE^y_i \bz_{t'} + c_i
    \right),
\end{align*}
where $\bE^y_i$ is the $i$-th row vector of the target word embedding matrix.

The neural machine translation model is usually trained to maximize the
log-probability of the correct translation given a source sentence using a large
training parallel corpus. This is done by stochastic gradient descent, where the
gradient of the log-likelihood is efficiently computed by the backpropagation
algorithm.

\section{Contextualized Word Embedding Vectors}

\subsection{Word Embedding Vectors}

One-hot representation of a word in Eq.~\eqref{eq:one_hot} is unique in the
sense that each and every word in a vocabulary is equally distant from every
other word. This implies that the words lose all the information relative to the
other words, when represented as a one-hot vector. The meaning of a word,
relative to those of the other words in the vocabulary, is thus learned through the
associated word embedding vector
(Eqs.~\eqref{eq:word_emb_x}--\eqref{eq:word_emb_y}) during training. In other
words, training brings similar words close to each other in the word embedding
space and dissimilar words far away from each other.

This phenomenon of similarity learning via word embedding vectors has been
observed in many different natural language processing tasks done with neural
networks. Already in 1991, Miikkulainen~and~Dyer~\cite{Miikkulainen1991} noticed
that training a neural network with one-hot vectors as its input learns the word
embedding vectors that ``{\it code properties of the input elements that are
most crucial to the task}.'' Based on this observation
Bengio~et~al.~\cite{Bengio2003} proposed to build a neural network based
language model and found that it generalizes better to unseen or rare $n$-grams:
the word embedding vectors capture similarities between words and the neural
network can learn a smooth mapping that automatically generalizes by producing
similar outputs for semantically similar input sequences.  The interest in word
embedding vectors, or distributed representation of words, was fueled by the
earlier observations that these unsupervised word embedding vectors can be used
to improve supervised natural language tasks
greatly~\cite{collobert2011natural,turian2010word}. 

\subsection{Multiple Dimensions of Similarity}
\label{sec:polysemy}

An important characteristic of the high-dimensional word embedding vectors is
that it encodes multiple dimensions of similarities. This is necessary in order for
a neural network to cope with polysemy. We can qualitatively check this
phenomenon of multiple dimensions of similarities by inspecting a local chart of
the manifold on which the word embedding vectors reside. 

For any word $\bx'$ under inspection, we find the $N-1$ nearest neighbours
$\left\{ \bx^1, \ldots, \bx^{N-1} \right\} \subset \bE_x$ in the word embedding
matrix.  The $N$ word embedding vectors $\left\{ \bx', \bx^1, \ldots, \bx^{N-1}
\right\}$ now characterize a local chart centered at $\bx'$, and we use
principal component analysis (PCA) to find the corresponding lower-dimensional
Euclidean space. In this Euclidean space, we can inspect the nearest neighbours
along each coordinate.\footnote{
    The code for this analysis is available publicly at
    \url{https://github.com/kyunghyuncho/WordVectorManifold}.
}
In Table~\ref{tab:polysemy}, we show two such examples using the word embedding
vectors trained as a part of the continuous-bag-of-word (CBoW)
network~\cite{Mikolov2013}.\footnote{
    We used the word embedding vectors provided as a part of
    \cite{hill2015learning}. 
}
These examples clearly show that each word embedding
vector encodes more than one notions of similarities. A similar behaviour can
only be observed with multi-map t-SNE~\cite{van2012visualizing}.

\begin{table}[t]
    \centering
    \begin{tabular}{l | c || l}
        Word $\bx'$ & Axis & Nearest Neighbours \\
        \hline
        \hline
        \multirow{2}{*}{notebook} & 1 & diary notebooks (notebook)
        sketchbook jottings \\
        & 2 & palmtop notebooks (notebook) ipaq laptop \\
        \hline
        \multirow{2}{*}{power} & 1 & powers authority (power) powerbase
        sovereignity \\
        & 2 & powers electrohydraulic microwatts hydel (power)
    \end{tabular}
    \caption{The placements of the word embedding vectors along the principals
        axes in the local charts of ``notebook'' and ``power''. In the case of
        ``notebook'', it is clear that the first axis corresponds to different
        types of books for writing a note, while the second axis corresponds to
        portable computing devices. In the case of ``power'', the first axis
        corresponds to political or social authority, while the second axis to
        physical energy. 
    }
    \label{tab:polysemy}
\end{table}

\subsection{Contextualized Word Embeddings}
\label{sec:contextualization}

The fact that each word embedding vector represents multiple dimensions of
similarity implies that a subsequent part of a neural network needs to
disambiguate the word based on the context in which the word was used. In the
case of neural machine translation, we should consider source and target word
embedding vectors separately. The encoder, which is implemented as a bidirectional
recurrent network, can disambiguate it by using all the other words in a source
sentence. On the other hand, the decoder can exploit both the previous words in
a target sentence as well as all the source words. 

Already pointed out in 1949 by Weaver~\cite{weaver1949translation}, much of
ambiguity in word meaning can be resolved by considering surrounding words.
The consequence of this in neural machine translation is that the recurrent
network, either in the encoder or decoder, needs to remember all the previous
words until the word in question in order to decide its meaning disambiguously.
In other words, the encoder and decoder must sacrifice their capacity in
disambiguating the words. This is undesirable, as what we truly want the encoder
and decoder to do is to capture the higher-level compositional structures of a
sentence that are necessary for translation.

\begin{figure}[t]
    \hfill
    \begin{minipage}{.45\textwidth}
        \centering
        \includegraphics[height=2.5in]{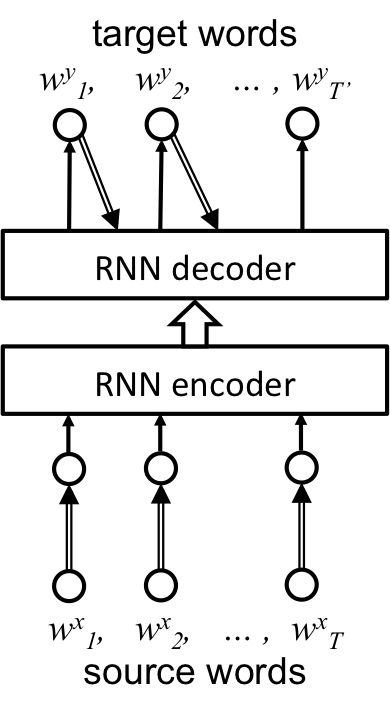}

        (a)
    \end{minipage}
    \begin{minipage}{.45\textwidth}
        \centering
        \includegraphics[height=2.5in]{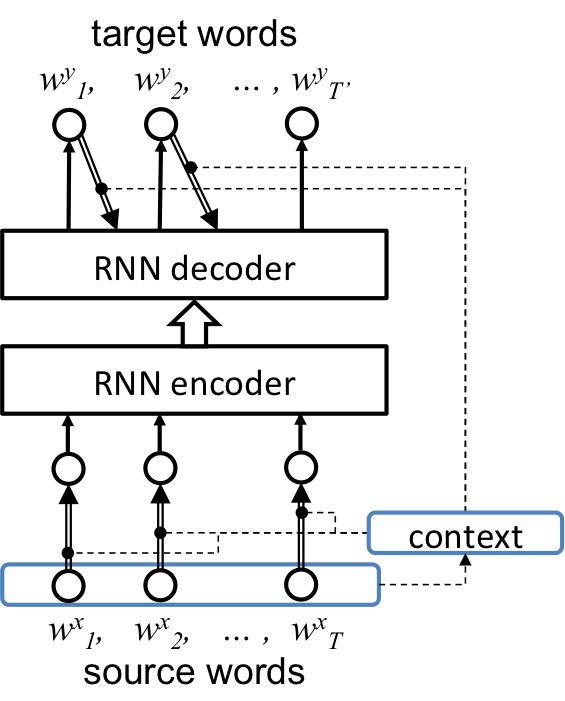}

        (b)
    \end{minipage}
    \hfill

    \caption{Graphical illustrations of the neural machine translation system
        (a) without and (b) with the proposed contextualization.}
\label{fig:nmt_network}
\end{figure}

In order to address this issue and to reduce the burden from the recurrent
networks in a neural machine translation system, we propose to {\it
contextualize} the word embedding vectors before being fed to the recurrent
networks as shown in Fig.~\ref{fig:nmt_network}. This contextualization disambiguates the word's meaning by {\it masking} out some dimensions of the word embedding vectors based on the context.

Let us describe in detail the proposed contextualization. First, we define the
context $\bc^x$ as a representation of the unordered set of all source words and compute it as the
average of the nonlinearly transformed source word embedding vectors, i.e.,
\begin{align}
    \label{eq:context_ext}
    \bc^x = \frac{1}{T} \sum_{t=1}^T \NN_{\TT}(\bx_t),
\end{align}
where $\NN_{\TT}:\RR^E \to \RR^C$ is a feedforward neural network parametrized
by $\TT$. 

Then, we compute a context mask from $\bc^x$ for each word embedding vector {\it
before} it is fed into an input or output recurrent network. This is simply done by
\begin{align*}
    \bx_t \leftarrow \bx_t \odot \sigma(\bW_x \bc^x + \bb_x), \\
    \by_t \leftarrow \by_t \odot \sigma(\bW_y \bc^x + \bb_y), 
\end{align*}
where  $\odot$ is an element-wise multiplication and $\sigma: \RR^C \to \left[
0, 1\right]$ is an element-wise sigmoid function and transforms the context to a
binary-like mask, with $\bW_x$ and $\bW_y$ (weight matrices), $\bb_x$ and $\bb_y$
(bias vectors) being additional parameters.

Recall that we introduced earlier in Sec.~\ref{sec:polysemy} a complicated
procedure based on estimating a local chart on a manifold using PCA. Compared to
that procedure, our proposal for contextualization may seem simple and perhaps
insufficient. This is however not a serious issue due to two reasons. First, the
context mask $\bc^x$ is extracted via a highly nonlinear function in
Eq.~\eqref{eq:context_ext} which can learn to effectively project axes from the
Euclidean space, corresponding to a local chart, back to the coordinates on the
manifold. Second, as both the word embedding vectors are learned together, we
expect the dimensions of the word embedding vector to disentangle multiple
dimensions of similarity automatically to maximize the use of the proposed
contextualization routine.  Although both of these are not easily verifiable, we
show later in the experiment that the proposed contextualization indeed improves
the translation quality.

\section{Symbolization}

\subsection{Proper Nouns, Digits and Rare Words}

The use of continuous vectors as an intermediate representation of source and
target sentences in neural machine translation greatly improves generalization
of machine translation by avoiding the issue of data sparsity (see, e.g., Sec.
5.2.1 in \cite{cho2015natural}.) This, however, brings in some unnecessary
complications as well. One such complication is in handling proper nouns,
digits and rare words.  

First, the rare words, which often occur up to a handful of times in a whole
training corpus, are problematic, because their word embedding vectors cannot be
well estimated during training. It is hence a usual practice to map all those
rare words whose frequencies are under a predefined threshold to a single token
representing an {\it unknown} word. This approach has been used without much
problem in language modeling, where the objective is to {\it score} a given
sentence. However, it is unacceptable for a machine translation system to
generate these {\it unknown} symbols when generating a translation.

The second issue is with digits.  Clearly the meaning of any digit is not
defined by its context, and there is a clear, single meaning associated with
each digit, that is the number denoted by it. The conversion to and from a
continuous vector space of a digit may therefore introduce unnecessary noise,
resulting in an incorrect translation. 

Proper nouns often exhibit both of the above problems. They are often rare,
except for the proper nouns of a famous person, organization, object or
location.  Furthermore, the transformation between a proper noun and a
high-dimensional continuous vector likely introduces noise that is absolutely
not necessary, as the meaning of such a proper noun is fixed to a single entity.

These issues and their underlying causes suggest that it would be beneficial to
treat these three cases separately from all the other words.  This is precisely
what we propose to do, and we will describe it in more detail below. 

\subsection{Previous Approaches}

In most of these special cases, a source word is directly copied to a target
sentence either as it is or after a simple transformation via dictionary lookup.
This property suggests a simple algorithm that can be run outside neural machine
translation. Assuming that there exists an alignment between a rare source word
and a word or placeholder in the target sentence, we can look up the rare word
in a pre-built dictionary and replace the target-side placeholder with the
queried word which can be the source word itself or its appropriate translation
or transliteration.

Based on this observation, Jean~et~al.~\cite{Jean2015} earlier proposed a number
of heuristics for handling rare words with the attention-based neural machine
translation. In their approach, the attention weights $\alpha_{t', t}$ from
Eq.~\eqref{eq:att_weight} are used to determine the source word aligned to each
of the unknown tokens in the generated translation, i.e., \mbox{$\argmax_t
\alpha_{t', t}$}. The source word determined by this mechanism is translated
word-wise by looking up a pre-built dictionary, and the corresponding unknown
token in the translation is replaced with the result of the look-up.

Simultaneously, Luong~et~al.~\cite{Luong2015} proposed another mechanism that
does not require the attention mechanism. They used an external alignment
mechanism, such as IBM~Model~2~\cite{brown1993mathematics}, to find the
alignment between the source and target words in the training corpus. This
alignment is used to assign multiple, numbered {\it unknown} tokens for rare
words so that the correspondences between these tokens in the source and target
sides are maintained. For instance, the first unknown token in the source side
and its corresponding target word will be replaced with
$\left<\text{UNK}_1\right>$, regardless of the target word's position in the
translation. During test time, given a source sentence and a generated
translation, each of the unknown tokens in the translation will be replaced by
querying the corresponding source-side word in a pre-built dictionary.

Both of these approaches have been highly effective in improving the translation
quality. For instance, Jean~et~al.~\cite{Jean2015} reported +3 and +2.5 BLEU
improvement with the simple replacement technique based on the attention
mechanism on En-Fr and En-De, respectively. Similarly,
Luong~et~al.~\cite{Luong2015} reported +1.6--+2.8 BLEU improvement on En-Fr with
their approach based on the positional unknown tokens
$\left<\text{UNK}_n\right>$.

\subsection{Symbolization of Proper Nouns, Digits and Rare Words}

In this paper, we extend the approach by Luong~et~al.~\cite{Luong2015}, which is
based on the positional unknown tokens, to include multiple positional special
tokens. Instead of a single special token \UNKS{n}, we propose to use three
special tokens:
\begin{enumerate}
    \item \NSym{n}: Digit
    \item \SSym{n}: Proper noun
    \item \CSym{n}: Acronym
\end{enumerate}
See Table~\ref{table:symbol} for examples of symbolization.

\paragraph{Digit \NSym{n}} 
The basic idea is to replace any consecutive digits (without blank spaces
in-between) appearing both in the source and target sentences with the special
symbol $\NSym{n}$, where $n$ denotes its order in the source sentence. In order
to better address the cases of a number followed immediately by its unit, we
separate the unit or any non-digit characters from consecutive digits (e.g.,
`137Kg' $\to$ `137', `Kg'.) Also, we normalize the variations in writing a long
digit such as decimal marks (`,' vs. `.') and digit grouping deliminators (`,'
vs. `.' vs. ` ',) when matching digits in the source and target sides.

\paragraph{Proper Noun \SSym{n}} 
In many of the European languages, which we mainly consider in this paper,
capitalization is used to indicate that a word is a proper noun. Therefore, we
replace a maximal consecutive phrase of more than one capitalized words
appearing on both the source and target side, with the special symbol
$\SSym{n}$. Similarly to $\NSym{n}$, $n$ denotes its order in the source
sentence. We consider a phrase, as many proper nouns are often proper phrases
(e.g., `New York'.) When spotting these proper noun phrases, we do not consider
non-capitalized functional words such as `of' in order to properly handle noun
phrases in the form of `X of Y' (e.g., `World of Warcraft'.)

\paragraph{Acronym \CSym{n}} 
Most of the acronyms can be handled similarly to proper nouns, except that there
are cases where acronyms of a single entity differ across languages. For
instance, the acronym of `International Monetary Fund' is `IMF' in English but
`FMI' in French. We notice that it is rare to have more than one such cases in a
single sentence pair. We first replace all the matching acronym pairs. If there
is only one all-capital word left in each of the source and target sentences, we
consider them a match and replace it with $\CSym{n}$.

\paragraph{Rule Dictionary}
We construct a mapping rule for each matching pair that was replaced by one of
the special symbols. This rule dictionary is used during test to replace the
generated special symbols in a translation. Although it is certainly possible to
incorporate external rules into this dictionary, we do not test it to avoid
including any external resource when evaluating the proposed symbolization
technique.

Note that this approach of symbolization has been used to a certain extent in
more conventional statistical machine translation and language modelling. For
more discussion, we refer the reader to Sec.~7.4 of \cite{koehn2010statistical}.
This approach however has not been adopted widely in neural machine translation
yet.

\begin{table}[t]
\begin{center}
\begin{tabular}{| l | p{4.6in} |}
\hline
En & \small The \underline{World of Warcraft} and \underline{Warcraft III}
American regional finals were held at the \underline{House of Blues} on June
\underline{2}nd and \underline{3}rd in \underline{San Diego} , California . \\
\hline
Fr & \small Les finales régionales de \underline{Warcraft III} et de
\underline{World of Warcraft} pour l\&apos; Amérique du Nord se sont
déroulées au \underline{House of Blues} , les \underline{2} et
\underline{3} juin derniers à \underline{San Diego} . \\
\hline
\hline
En & \small The \SSym{1} and \SSym{2} American regional finals were held at the
\SSym{3} on June \NSym{1} \^{}\^{}nd and \NSym{2} \^{}\^{}rd in \SSym{4} , California
.\\
\hline
Fr & \small Les finales régionales de \SSym{2} et de \SSym{1} pour l\&apos;
Amérique du Nord se sont déroulées au \SSym{3} , les \NSym{1} et
\NSym{2}
juin derniers à \SSym{4} .\\
\hline
\end{tabular}
\end{center}
    \caption{Examples of symbolization applied to En-Fr sentence pairs. \^{}\^{}
    is a special indicator that the words immediately before and after be merged
without a blank space after de-symbolization.}
\label{table:symbol}
\end{table}

\section{Experimental Settings}

\subsection{Tasks and Corpora}

We evaluate the neural machine translation systems with and without the proposed
methods on two tasks of English-to-French (En-Fr) and English-to-German (En-De)
translation. We use all the parallel corpora made publicly available via
WMT'14.\footnote{
    \url{http://www.statmt.org/wmt14/}
} 
The En-Fr corpora are cleaned following the procedure in \cite{cho2014learning},
and after tokenization, has approximately 250M words (English side.) The En-De
corpora are prepared following the procedure in \cite{Jean2015}, resulting in
approximately 100M words (English side.)  

We use newstest-2014 and newstest-2013 as the development sets and
newsdiscusstest2015 and newstest2015 as the test sets for En-Fr and En-De,
respectively. 

\subsection{Vocabulary Preparation}
\label{sec:vocab}

Both of the corpora go through a minimal set of preprocessing. First, we
tokenize them using the script provided as a part of Moses.\footnote{
    \url{https://github.com/moses-smt/mosesdecoder}
} As a comparison, we evaluate a setting where byte pair encoding (BPE) is used,
as a replacement of the proposed symbolization, to extract sub-word
symbols~\cite{Sennrich2015}. This has been found to be an effective and
efficient approach to handling the issue of a large target
vocabulary~\cite{sennrich2015improving,firat2016multi,chung2016character} and is
known to be able to transliterate rare, proper nouns up to a certain
extent~\cite{Sennrich2015}. In
all the cases, we use up to top-30k most frequent symbols (either tokenized
words or BPE-based sub-words.) 

The proposed symbolization is applied right after the initial tokenization. In
the case of using BPE, the BPE segmentation is done on the symbolized corpus,
and in the test time, the BPE de-segmentation is followed by the
de-symbolization.

\begin{table}[t]
\begin{center}
\begin{tabular}{|l|c | c |c | c|}
\hline
{\bf En-Fr} & \multicolumn{2}{c|}{English} & \multicolumn{2}{c|}{French} \\
Symbolization & Before & After & Before & After \\
\hline
Unique words & 1.77M & 0.44M &  1.73M & 0.49M \\
Total words & 294M & 258M & 336M  & 297M  \\
Coverage (\%) & 96.1 & 99.1 & 96.1 & 98.9\\
\hline
\hline
{\bf En-De} & \multicolumn{2}{c|}{English} & \multicolumn{2}{c|}{German} \\
Symbolization & Before & After & Before & After \\
\hline
Unique words & 709K & 411K &  1,553K & 1,035K \\
Total words & 117M & 102M & 110M  & 95.5M  \\
Coverage (\%) & 97.7 & 98.6 & 93.3 & 95.3 \\
\hline
\end{tabular}
\end{center}
    \caption{Corpora Statistics}
\label{table:voc}
\end{table}%

Aside the BPE segmentation, we noticed that the number of unique tokens on each
corpus greatly decreases after the proposed symbolization is applied. This
improves the coverage by the 30k-word-large vocabulary as much as 3 percentage
points. See Table~\ref{table:voc} for detailed statistics.

\subsection{Training}

We use the very same attention-based neural translation model from
\cite{Bahdanau2015}. The only change we make is to use long short-term memory
units (LSTM) instead of gated recurrent units (GRU). Further, instead of
Adadelta~\cite{zeiler2012adadelta}, we use Adam~\cite{Kingma2014adam} for
adaptive learning rate adjustment with gradient clipping (threshold at $1$.) As
the proposed contextualization and symbolization do not alter the internals of
the neural machine translation model, we use this model and training
configuration for all the experiments.

We remove any sentence pair if more than 10\% and 30\% of all the words are
out-of-vocabulary after the symbolization for En-Fr and En-De, respectively.
Also, we only use sentence pairs of which both sentences are only up to 50
symbols long. We early-stop any training when the BLEU score on a development
set does not improve any more. 

\subsection{Evaluation}

A simple forward beamsearch is used to approximately find the most likely
translation given a source sentence from a trained model. We set the width of
the beam to 12 following \cite{Bahdanau2015}. The generated translations are
scored against the reference translations using `multi-bleu.perl' script from
Moses.\footnote{
    \url{https://github.com/moses-smt/mosesdecoder/blob/master/scripts/generic/multi-bleu.perl}
} This evaluation is done on {\it tokenized} sentences.

\section{Result and Analysis}

\subsection{Quantitative Analysis}

In Table~\ref{table:bleu}, we present the translation qualities, measured by
BLEU on the test sets, of all the trained models on both En-Fr and En-De. The
most noticeable observation is that both of the proposed
methods--contextualization and symbolization-- improve the translation quality
over the baseline model which is a vanilla attention-based neural translation
system (+Context and +Symbol.) Furthermore, the proposed contextualization and
symbolization are complementary to each other, and the most improvement is made
when both of them are used together (Baseline+Context+Symbol.)

As expected, the proposed symbolization has a similar effect as the BPE
segmentation does (Baseline+Symbol vs. Baseline+BPE) on both of the language
pairs. We observe that the proposed contextualization, which was found to be
complementary to the symbolization, improves the translation quality even when
the BPE-base subword symbols were used. 
%Furthermore, even when the BPE-based
%subword symbols are used, using the contextualization and symbolization together
%helps the neural machine translation system achieve the best translation
%quality (Baseline+BPE+C+S).

\begin{table}[t]
    \centering
\begin{tabular}{l||c|c|c|c|c|c}
& \multicolumn{2}{c|}{En-Fr} & \multicolumn{2}{c|}{En-De} \\
\hline
Beam Width 	&  1 & 12 			& 1 & 12 \\ 
\hline
\hline
\small Baseline		& 24.54 {\scriptsize (27.02)} & 27.62 {\scriptsize (29.21)}
& 15.01 {\scriptsize (15.84)}& 15.93 {\scriptsize (17.05)}\\
\hline
\small +Context (C)	& 27.62 {\scriptsize (29.01)}& 28.87 {\scriptsize (30.72)}	&16.24 {\scriptsize (16.68)} & 17.82 {\scriptsize (17.89)}\\
\small +Symbol (S)	& 27.05 {\scriptsize (29.51)}& 29.39 {\scriptsize (31.83)}	&17.78 {\scriptsize (17.55)} & 19.45 {\scriptsize (19.00)}\\
+C+S		&         28.58 {\scriptsize (31.22)}& {\bf 30.10} {\scriptsize (33.30)}  &18.87 {\scriptsize (18.34)} & {\bf 20.79} {\scriptsize (19.55)}\\
\hline
\small +BPE		     & 27.05 {\scriptsize (29.10)}& 30.35 {\scriptsize (32.32)}	    & 17.43 {\scriptsize (17.51)}& 20.51 {\scriptsize (19.97)}\\
\small +BPE+C        & 29.65 {\scriptsize (31.39)}& {\bf 32.02} {\scriptsize (33.80)}	    & 19.09 {\scriptsize (18.84)}& 21.95 {\scriptsize (20.81)}\\
\small +BPE+C+S      & 29.32 {\scriptsize (31.33)}& 31.19 {\scriptsize (33.83)}       & 19.54 {\scriptsize (19.48)}& {\bf 21.99} {\scriptsize (20.97)}
\end{tabular}
\caption{BLEU scores on the test sets for En-Fr and En-De with two different
beam widths. The scores on the development sets are in the parentheses.}
\label{table:bleu}
\end{table}%

\subsection{Effect of Contextualization}

We conjectured earlier in Sec.~\ref{sec:contextualization} that the
contextualization helps translation by selectively masking out irrelevant some
dimensions of word meaning based on the context. This directly implies that the
neural translation system with the contextualization will be less confused among
many similar words. Browsing through the translations of the source sentences
included in the development and test sets, we have observed many such cases, and
the following is one such example:
\begin{quote}
    \it
    The Scorsese-DiCaprio duo seems to have
    \colorbox{purple!20}{rediscovered} the magic that brought them
    together in Shutter Island .
\end{quote}

The above sentence is translated by the neural translation system with BPE into
\begin{quote}
    \it
    Le duo de Corsso-DiCaprio semble avoir \colorbox{purple!20}{retrouvé} la
    magie qui les a réunis dans l'île Shuttle . 
\end{quote}
Meanwhile, the same sentence is translated into
\begin{quote}
    \it
    Le duo Scorson-DiCaprio semble avoir \colorbox{purple!20}{redécouvert} la
    magie qui les rassemblait dans l\&apos; île de Shutter .
\end{quote}
when the contextualization was used.

Without the contextualization, the word \colorbox{purple!20}{rediscovered} was
translated into ``retrouv\'e'' (``found'' according to Collins French-English
Dictionary\footnote{
    \url{http://www.collinsdictionary.com/dictionary/french-english/retrouver}
}). This is not an incorrect translation, but the translation of this word into
``red\'ecouvert'' (``rediscovered'' according to the same source\footnote{
    \url{http://www.collinsdictionary.com/dictionary/french-english/red\%C3\%A9couvrir}
}) by the contextualized model is closer to the original English word. This is
an example of the source-side context disambiguating the target word. 

Let us provide another example in the case of En-De. The following source sentence  
\begin{quote}
    \it
    There are \colorbox{blue!20}{13,435} cumulative cases (12,158 AIDS and 1,317 HIV).
\end{quote}
is translated by the neural translation model with BPE into 
\begin{quote}
    \it 
    Es gibt \colorbox{blue!20}{13.335} kumulierte Fälle (12,158 Aidg und 1,317 HIV).
\end{quote}
The same sentence is translated by the same model with the contextualization
into
\begin{quote}
    \it
    Es gibt \colorbox{blue!20}{13.435} kumulative Fälle (12.158 AIDS und 1.317 HIV).
\end{quote}

Notice that the tokens \colorbox{blue!20}{13,435} and \colorbox{blue!20}{13.335}
are close to each other in the word embedding space, as they represent close-by
numbers. This makes it difficult for a vanilla neural machine translation model
to distinguish between those two. The contextualization, as seen above, can
successfully disambiguate ``13.435'' from ``13.335'' and correctly puts the
former. 

\subsection{Effect of Symbolization}

As briefly mentioned earlier in Sec.~\ref{sec:vocab}, it is known that neural
machine translation transliterates rare (proper) nouns using BPE-based subword
symbols. This is however not perfect, as the rules for transliteration exhibit
many exceptions, and statistical generalization often fails to account for rare
proper nouns. For instance, the following sentence
\begin{quote}
    \it
    \colorbox{red!20}{Trakhtenberg} was the presenter of many programs before
    \colorbox{green!30}{Hali-Gali} times.
\end{quote}
is translated by the model using BPE-based subword symbols into
\begin{quote}
    \it
    \colorbox{red!20}{Trakttenberg} war der Moderator vieler Programme vor
    \colorbox{green!30}{Hali-Gi} Male.
\end{quote}
It is easy to notice that the model has failed at correctly transliterating
``Trakhtenberg''. On the contrary, once the symbolization is used (together with
the contextualization), the model correctly translates the source sentence above
into
\begin{quote}
    \it
    \colorbox{red!20}{Trakhtenberg} war der Moderator von vielen Programmen vor
    \colorbox{green!30}{Hali-Gali}-Zeiten.
\end{quote}
A similar behaviour is also observed with ``Hali-Gali''.  This example clearly
demonstrates the effectiveness of the proposed symbolization, even when the
BPE-based subword symbols are used.

The advantage of the symbolization is more apparent when words are used as basic
symbols instead of BPE-based subword symbols. The same source sentence is
translated into
\begin{quote}
    \it 
    \colorbox{red!20}{UNK} war der Moderator von vielen Programmen vor
    \colorbox{green!30}{UNK}.
\end{quote}
using the word-based translation model with the contextualization only. 
By including the symbolization as pre- and post-processing routines, we get
\begin{quote}
    \it
    \colorbox{red!20}{Trakhtenberg} war der Moderator vieler Programme , bevor
    \colorbox{green!30}{Hali-Gali}-Zeiten UNK .
\end{quote}
It is clear that both of those proper nouns are correctly translated.

\section{Conclusions}

In this paper, we discussed the shortcoming of the existing neural machine
translation in terms of how a word or a symbol is represented as a continuous
vector and fed to recurrent networks. Based on the observation that a word
embedding vector encodes multiple dimensions of similarities, we proposed to
contextualize it by adaptively masking out each dimension of the source and
target embedding vectors based on the context of a source sentence.  In addition
to the contextualization of the word embedding vectors, we also propose to
symbolize special tokens to facilitate translating those tokens that are not
well-suited for translating via continuous vectors. 

The experiments on En-Fr and En-De revealed that both the contextualization and
symbolization techniques improve the translation quality of neural machine
translation significantly. Furthermore, we confirmed that these approaches are
agnostic to the type of linguistic symbols used to represent source and target
sentences by using byte-pair encoding to segment source and target sentences.
Especially, the proposed contextualization was found to be complementary to the
use of BPE-based subword symbols, and we find it an interesting future work to
test the contextualization with character-level neural machine
translation~\cite{ling2015character,chung2016character}.

\section*{Acknowledgments}
The authors would like to thank the developers of Theano~\cite{Bastien2012}.  We
acknowledge the support of the following agencies for research funding and
computing support: NSERC, Calcul Qu\'{e}bec, Compute Canada, the Canada Research
Chairs, CIFAR and Samsung. KC thanks the support by Facebook and Google (Google
Faculty Award 2016).

\bibliography{hchoi_csl2016}
\bibliographystyle{jponew}

%\appendix

%\section{Supplemental Material}

\end{document}